\documentclass[sigconf]{acmart}
\AtBeginDocument{%
  \providecommand\BibTeX{{%
    \normalfont B\kern-0.5em{\scshape i\kern-0.25em b}\kern-0.8em\TeX}}}

\copyrightyear{2022}
\acmYear{2022}
\setcopyright{acmcopyright}\acmConference[SIGSPATIAL '22]{The 30th International Conference on Advances in Geographic Information Systems}{November 1--4, 2022}{Seattle, WA, USA}
\acmBooktitle{The 30th International Conference on Advances in Geographic Information Systems (SIGSPATIAL '22), November 1--4, 2022, Seattle, WA, USA}
\acmPrice{15.00}
\acmDOI{10.1145/3557915.3561026}
\acmISBN{978-1-4503-9529-8/22/11}

\usepackage{booktabs,multirow}
\usepackage{subfigure}
\usepackage{graphicx}
\usepackage{amsmath}
\usepackage{amsfonts}

\begin{document}

\newcommand{\ie}{\textit{i.e.}}
\newcommand{\eg}{\textit{e.g.}}
\newcommand{\name}{\text{AuxMobLCast}}
\title{Leveraging Language Foundation Models for \\ Human Mobility Forecasting}

\author{Hao Xue}
\authornote{Both authors contributed equally to this research.}
\affiliation{%
  \institution{University of New South Wales}
  \city{Sydney}
  \state{NSW}
  \country{Australia}
}
\email{hao.xue1@unsw.edu.au}
\author{Bhanu Prakash Voutharoja}
\authornotemark[1]
\affiliation{%
  \institution{University of Wollongong}
  \institution{University of New South Wales}
  \state{NSW}
  \country{Australia}
}
\email{bpv991@uowmail.edu.au}
\author{Flora D. Salim}
\affiliation{%
  \institution{University of New South Wales}
  \city{Sydney}
  \state{NSW}
  \country{Australia}
}


\begin{abstract}
In this paper, we propose a novel pipeline that leverages language foundation models for temporal sequential pattern mining, such as for human mobility forecasting tasks. For example, in the task of predicting Place-of-Interest (POI) customer flows, typically the number of visits is extracted from historical logs, and only the numerical data are used to predict visitor flows. In this research, we perform the forecasting task directly on the natural language input that includes all kinds of information such as numerical values and contextual semantic information. Specific prompts are introduced to transform numerical temporal sequences into sentences so that existing language models can be directly applied. We design an AuxMobLCast pipeline for predicting the number of visitors in each POI, integrating an auxiliary POI category classification task with the encoder-decoder architecture. This research provides empirical evidence of the effectiveness of the proposed AuxMobLCast pipeline to discover sequential patterns in mobility forecasting tasks. The results, evaluated on three real-world datasets, demonstrate that pre-trained language foundation models also have good performance in forecasting temporal sequences. This study could provide visionary insights and lead to new research directions for predicting human mobility.
\end{abstract}



\maketitle

\section{Introduction}
\begin{figure}
    \centering
    \includegraphics[width=.45\textwidth]{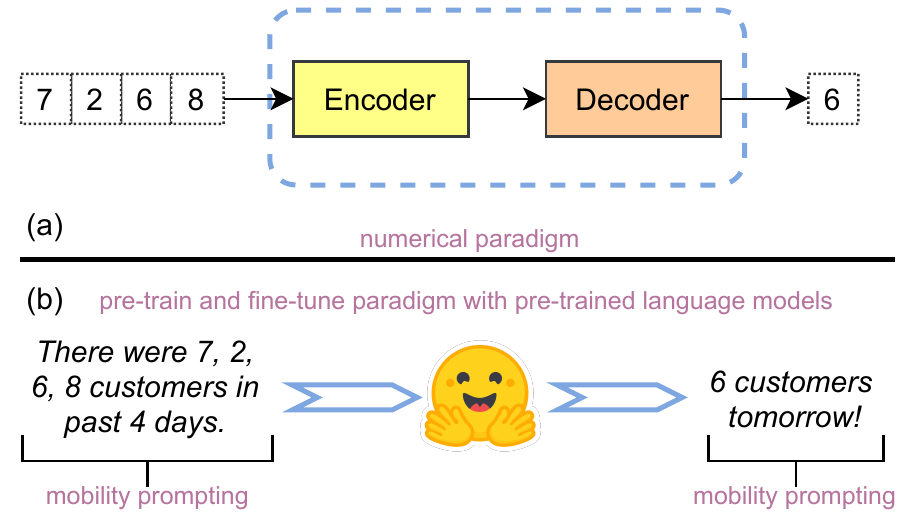}
    \caption{Conceptual comparison of: (a) the numerical paradigm for human mobility forecasting and (b) the proposed language foundation model based forecasting with mobility prompts.}
    \label{fig:intro}
\end{figure}

Nowadays, AI-powered digital assistants (\eg, Alexa and Siri) have demonstrated advanced performance in answering general topics whereas they still could not well-responding human mobility forecasting problems.
Mining spatio-temporal sequential patterns plays a critical component in many real-world applications including intelligent transportation and smart cities such as predicting the future customer flow of a shop to avoid the crowds during the pandemic.
We observe that almost all of the existing deep learning based solutions for spatio-temporal and human mobility prediction tasks can be summarised as a \textit{numerical paradigm} (Figure\ref{fig:intro} (a)) which takes a sequence of numerical values (history mobility observations) as input to generate a numerical value as future prediction (or a sequence of numerical values). 
Usually, only the numeric data can be well extracted and modelled within such a framework for future prediction.
Furthermore, this \textit{numerical paradigm} makes it difficult to seamlessly integrate the forecasting ability with the natural language processing models of the digital assistants.

More recently, the fast evolution of foundation models~\cite{bommasani2021opportunities} such as BERT~\cite{bert} and CLIP~\cite{clip} has led to a paradigm shift in designing, training, and applying deep learning models. In the new \textit{pre-train and fine-tune paradigm}, a foundation model is pre-trained with large-scale data and then adapted to solve various downstream tasks such as a well pre-trained BERT model for language translation, sentiment analysis, and question answering. However, in the literature, this shift only appears in Natural Language Processing (NLP) and Computer Vision (CV) fields. How to apply a foundation model for spatio-temporal forecasting and human mobility prediction remains unexplored.
In the time-series data forecasting domain especially with the human mobility data, due to the sequential numerical data format, there is no existing work on directly using pre-trained language foundation models for human mobility prediction.
Although some network architecture designs such as Transformer~\cite{vaswani2017attention} can be tweaked to the numerical paradigm for temporal sequence prediction (\eg, \cite{termcast}), it seems impossible to directly take advantage of existing pre-trained foundations models. 
The main reason is that the temporal sequence mining tasks require the prediction model to handle the numerical data format, whereas existing language foundation models are pre-trained with natural language sentences. This leads to a gap between the learned knowledge in pre-training and the downstream prediction.

In this paper, motivated by the above observations, we aim to investigate the following research question: \textit{how can we directly leverage existing pre-trained language foundation models for sequential temporal data?}
We hope that the exploration of this question in this vision paper could open new research directions and present novel insights in the spatio-temporal data forecasting field.
The mining of temporal sequential patterns is exemplified by an aggregated human mobility forecasting task in this work.
To address the aforementioned gap, we explore the idea of \textit{mobility prompting}, which transfers the human mobility data into natural language sentences so that the pre-trained language foundation models can be directly applied in the fine-tuning phase for predicting human mobility.
With mobility prompts that describe the historical mobility observations as input, we apply an encoder-decoder architecture to yield future mobility predictions (this forecasting paradigm is demonstrated in Figure\ref{fig:intro} (b)).
Compared to the numerical paradigm, directly using natural language as input can  seamlessly and naturally model the numerical sequences, the textual context, and the semantic information for generating future predictions.
Furthermore, we propose a novel~\name\ pipeline (\textbf{Aux}iliary \textbf{Mob}ility \textbf{L}anguage Fore\textbf{cast}ing) for forecasting human mobility, which introduces an auxiliary POI category classification task.
The auxiliary task is specially designed to associate with the {\fontfamily{qcr}\selectfont [CLS]} token in the encoder part (\eg, BERT's {\fontfamily{qcr}\selectfont [CLS]} token).
This extra customised task does not modify the foundation model structure which makes it possible to directly utilise existing pre-trained models (\eg, pre-trained weights provided by HuggingFace\footnote{https://huggingface.co/models}).
In particular, we conduct an empirical study to analyse the mobility forecasting performance of using different foundation models (such as BERT, RoBERTa, GPT-2, and XLNet) as the encoder/decoder of \name.
The experimental results demonstrate that \name\ can further improve the prediction performance.
In summary, our contributions are: 
\begin{itemize}
    \item 
    To the best of our knowledge, this paper presents the first attempt to fine-tune existing pre-trained language foundation models for forecasting human mobility data. Mobility prompts are specifically used to address the gap between the human mobility data and the language formats.
    \item 
    We empirically investigate and examine the performance of multiple pre-trained language foundation models for forecasting human mobility in the pre-train and fine-tune paradigm fashion. This work is the first demonstration of the power of pre-trained language models for forecasting tasks.
    \item We further introduce an auxiliary POI category classification task to improve the forecasting performance.
\end{itemize}

The rest of the paper is organised as follows. Section~\ref{sec:related} introduces the related work. Section~\ref{sec:model} presents a formal definition of the problem focused in this paper and describes the proposed \name. The details of datasets and our experimental settings are given in Section~\ref{sec:exp}. It also analysis the performance of our \name\ in human mobility forecasting and conducts ablation studies. Section~\ref{sec:conclusion} concludes the paper and discusses the potential impact and future directions of this work.

\section{Related Work}\label{sec:related}
\subsection{Language Foundation Models}
Recent rapid advances in self-supervised training techniques and Transformer~\cite{vaswani2017attention} architectures facilitate the emerging foundation models and the pre-train and fine-tune paradigm in various NLP tasks.
For example, BERT~\cite{bert} is a Transformer-based model pre-trained in a self-supervised fashion.
It utilises the masked language modelling task for pre-training on the unlabelled large-scale corpus of English text data, which makes the pre-trained model more scalable and can be adapted to different downstream tasks based on learned language understanding ability.
In the masked language modelling pre-training process, the model randomly masks a certain percentage (\eg, 15\%) of the words in the input sentence. The model is then trained to predict the masked words. 
Through this technique, the model can be pre-trained on the raw texts only without any human labelling required.
Since the emergence of BERT, almost all state-of-the-art models in NLP tasks are adapted and tweaked from one of a few foundation models such as BERT, GPT-2~\cite{gpt2} and RoBERTa.
Beyond the NLP field, the foundation models, as well as the pre-train and fine-tune paradigm, also demonstrate superior performance in image classification~\cite{clip,zhou2021learning} and speech recognition~\cite{baevski2020effectiveness}.
This work aims to investigate how to apply foundation models for mining temporal sequential patterns, which further broadens the application of the pre-train and fine-tune paradigm.
Although a similar numerical prediction task has been addressed with language models in \cite{spithourakis2018numeracy} and \cite{berg2020empirical}, our work differs from this task in that we are interested in sequential numbers instead of separate numbers in the input text.

\subsection{Time-series Forecasting}
The human mobility forecasting task is much related to general time-series forecasting research.
Deep learning based methods for time-series forecasting are mainly based on the Recurrent Neural Network (RNN) and its variants like Long Short Term Memory (LSTM) networks~\cite{hochreiter1997long} and Gated Recurrent Units (GRU)~\cite{chung2014empirical}.
Methods in this category often generate the future prediction in a sequence-to-sequence manner with the numerical history observations as the input sequence. 
Representative works under this numerical paradigm for sequential human behaviour prediction include ST-RNN~\cite{liu2016predicting} and DeepMove~\cite{feng2018deepmove}.

\begin{figure*}
    \centering
    \includegraphics[width=.875\textwidth]{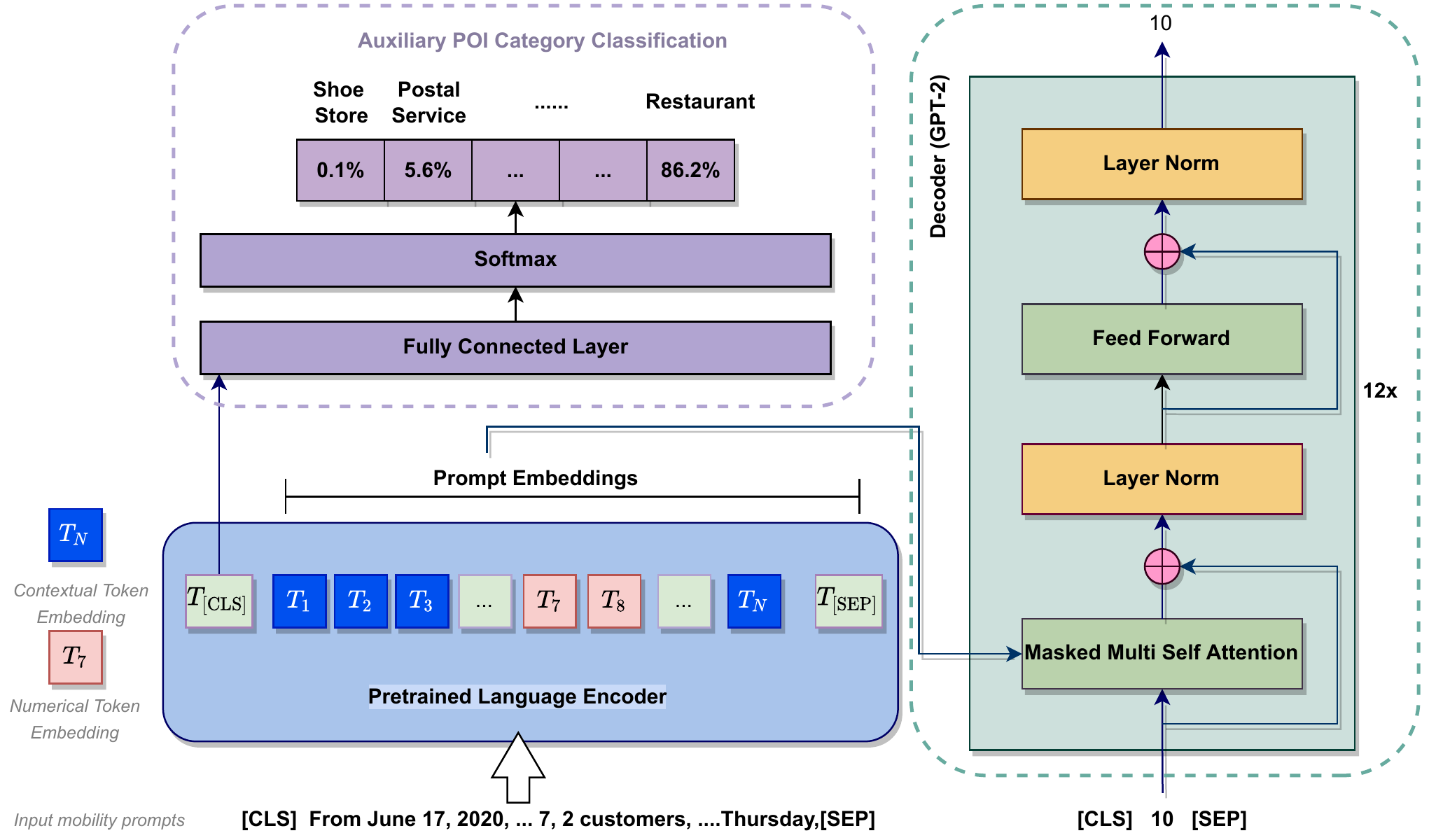}
    \caption{The illustration of our \name. Through the self-attention based language encoder, the interactions between the numerical tokens (red) and contextual tokens (blue, \eg, the temporal information) are simultaneously extracted in prompt embeddings for the decoder. We also explicitly introduce an auxiliary POI category classification (the purple part).}
    \label{fig:framework}
\end{figure*}

More recently, inspired by the success of Transformer~\cite{vaswani2017attention} in language processing, various methods such as~\cite{li2019enhancing,wu2020hierarchically, xue2021mobtcast, zhou2021informer} have been proposed for sequential time-series forecasting and human mobility prediction tasks.
Although these methods are based on the effective Transformer, they cannot fully take advantage of the pre-train and fine-tune paradigm due to two main reasons.
First, due to issues like privacy concerns in collecting human mobility data, there are no large scale time-series datasets that can be used for pre-training the model to learn general features of different time-series data types.
Second, these methods introduce unique designs and tweaks at the network structure level for modelling the time-series characteristics, which results in that existing pre-trained language models cannot be directly applied.

In this paper, we hypothesise that language models pre-trained with very large scale text data could also be beneficial in learning sequential patterns.
Instead of focusing on tweaking the model structure, we switch our focus to transforming the sequential data (\ie, aggregated human mobility in this work) into language descriptions through mobility prompts.
As a consequence, pre-trained language foundation models can be directly leveraged for forecasting human mobility.

\section{Methodology}\label{sec:model}
\subsection{Problem Definition}
In this paper, we focus on the aggregated human mobility forecasting task.
Let $\{\text{POI}_1 ,\text{POI}_2, \cdots, \text{POI}_M\}$ denotes a set of $M$ POIs in a certain area (\eg, a city).
For each $\text{POI}_m$, we can observe a history records of customer visits on $n$ continuous days: $x^m_{t_1: t_{n}}=[x^m_{t_1}, x^m_{t_2}, \cdots, x^m_{t_n}]$ where $x^m_t$ stands for the number of visits of POI $m$ on day $t$.
In addition, each POI corresponds to a semantic category class $c^m$ such as a \textit{Shoe shop} or a \textit{Post office}.
The focused human mobility forecasting problem can then be formulated as predicting the number of visits $x^{m}_{t_{n+1}}$ of the next day $t_{n+1}$ given the history observation $x^m_{t_1: t_{n}}$.

In this paper, we are particularly interested in generating the future $x^{m}_{t_{n+1}}$ under the \textit{pre-train and fine-tune paradigm} with language foundation models.
For this purpose, we propose to develop mobility prompt $X$ which translates the numerical history observations as natural language sentences instead of using the numerical sequence $x^m_{t_1: t_{n}}$ as the inputs of forecasting models in existing \textit{numerical paradigm} based time-series forecasting methods.
We note that the superscript $m$ (POI index) is dropped for simplification from hereon.

\subsection{Mobility Prompting}

Inspired by other work that apply language models for non-NLP tasks through prompting such as CLIP~\cite{clip} and CoOp~\cite{zhou2021learning}, we introduce mobility prompting to convert the sequential observation $x_{t_1: t_{n}}$ into language description $X$ to leverage pre-trained language models for forecasting human mobility.
Such a transformation step is a key enabler for taking advantage of the \textit{pre-train and fine-tune paradigm} in this work.
The purpose of mobility prompting is to pre-process
the numerical mobility data to generate meaningful sentences that can be fed to the pre-trained language foundation models.
Specifically, we develop and investigate three different prompts as illustrated in Table\ref{tab:prompt}.
In general, all three types of prompts contain key components related to mobility in the input text part, including the mobility data (\ie, the number of visits on each day, the red parts), the temporal information (\ie, the date, the blue parts), and information about the POI itself (the orange parts).
The temporal information includes not only the history observations timestamps (\eg, {\fontfamily{qcr}\selectfont From June 17, 2020, Wednesday to July 01, 2020, Wednesday}) but also the time of the prediction target (\eg, {\fontfamily{qcr}\selectfont July 02, 2020, Thursday}) which provides an important cue for prediction.
For the output target text (used as the ground truth
for fine-tuning and evaluation), all prompts include the future prediction target $x_{t_{n+1}}$ (\eg, $11$ in the table).

\begin{table}[]
\centering
\caption{Examples of three different mobility prompt types used in this study. 
Mobility data, temporal information, and the POI information are given in \textcolor{red}{red}, \textcolor{blue}{blue}, and \textcolor{orange}{orange}, respectively.}
\begin{tabular}{l|p{2.75in}} 
\hline   \hline
\multirow{2}{*}{\rotatebox[origin=c]{90}{\textit{Prompt A~}}} &
  \textit{\textbf{Input}}: “Place-of-Interest (POI) \textcolor{orange}{385} is a \textcolor{orange}{Limited-Service Restaurant}. From \textcolor{blue}{June 17, 2020, Wednesday to July 01, 2020, Wednesday}, there were \textcolor{red}{11, 11, 10, 12, 9, 12, 6, 13, 10, 15, 16, 8, 8, 13, 19} people visiting POI \textcolor{orange}{385} on each day. On \textcolor{blue}{July 02, 2020, Thursday},” \\
 & \textit{\textbf{Target}}: “there will be \textcolor{red}{11} people visiting POI \textcolor{orange}{385}.” \\ \hline \hline  
\multirow{2}{*}{\rotatebox[origin=c]{90}{\textit{Prompt B~}}} &
  \textit{\textbf{Input}}: “From \textcolor{blue}{June 17, 2020, Wednesday to July 01, 2020, Wednesday}, there were \textcolor{red}{11, 11, 10, 12, 9, 12, 6, 13, 10, 15, 16, 8, 8, 13, 19} people visiting POI \textcolor{orange}{Limited-Service Restaurant} on each day. On \textcolor{blue}{July 02, 2020, Thursday},” \\
 & \textit{\textbf{Target}}: “there will be \textcolor{red}{11} people.”                  \\ \hline \hline  
\multirow{3}{*}{\rotatebox[origin=c]{90}{\textit{Prompt C~}}} &
  \textit{\textbf{Input}}: “From \textcolor{blue}{June 17, 2020, Wednesday to July 01, 2020, Wednesday}, there were \textcolor{red}{11, 11, 10, 12, 9, 12, 6, 13, 10, 15, 16, 8, 8, 13, 19} people visiting POI on each day. On \textcolor{blue}{July 02, 2020, Thursday},” \\
 & \textit{\textbf{POI Category Target}}: “\textcolor{orange}{Limited-Service Restaurant}”    \\ 
 & \textit{\textbf{Mobility Target}}:  “\textcolor{red}{11}”    \\ \hline\hline  
\end{tabular}
\label{tab:prompt}
\end{table}

Table\ref{tab:prompt} lists an example input/target text for each prompt.
Prompt A is the basic format.
Compared to Prompt A, the POI index id is removed from the prompting sentences in Prompt B whereas the category information is kept.
There are two rationales for this alteration: (1) As demonstrated in~\cite{xue2021mobtcast}, POIs from different categories often have different visiting patterns. 
As a consequence, it is important to keep the category information.
(2) Excluding a specific POI id could further alleviate privacy concerns in real-world applications.
Based on Prompt B, we further design Prompt C for our \name\ pipeline. For this prompt type, the output target is explicitly divided into two parts: the POI category target for the auxiliary POI category classification in \name\ (details given in the next section) and the direct mobility prediction target.

\subsection{Proposed \name\ }

\begin{figure}
    \centering
    \includegraphics[width=.375\textwidth]{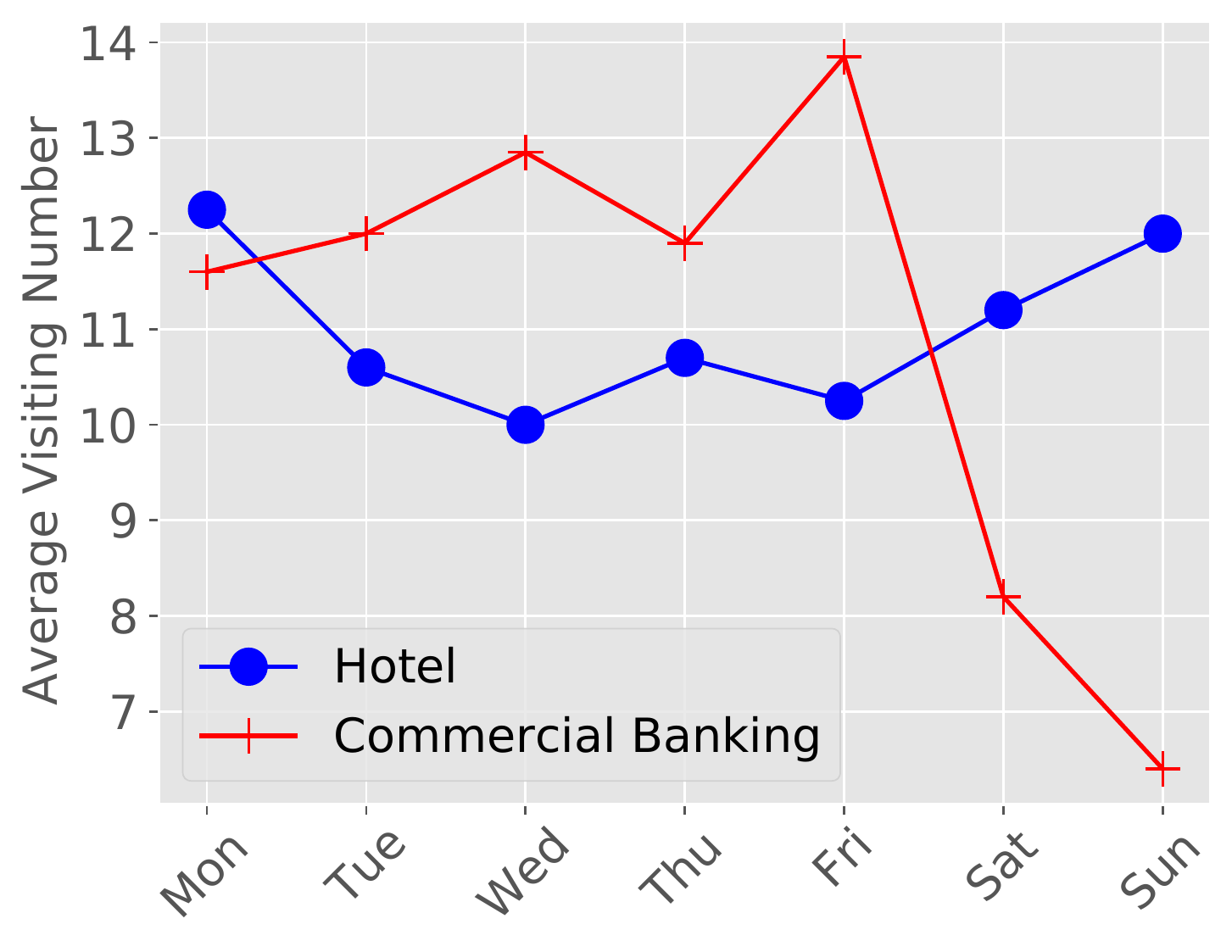}
    \caption{The average weekly visiting patterns of two different categories POIs.}
    \label{fig:category}
\end{figure}

In this study, in addition to investigating existing pre-trained language models' ability on human mobility forecasting, we also propose a novel pipeline \name\ based on the general encoder-decoder framework.
As demonstrated in Figure\ref{fig:framework}, in \name, we explicitly introduce an auxiliary classification task (the purple part) to classify the POI category.
This design is motivated by the observation that the POI category is correlated to its visiting pattern.
For example, Figure\ref{fig:category} shows the averaged weekly visiting pattern of two POIs located in Miami during the entire data collection period (details given in Section~\ref{sec:dataset}).
One POI belongs to the \textit{Hotel} category and the other POI is the \textit{Commercial Banking} category.
From this plot, we can clearly see that these two POIs have different visiting patterns. The Hotel POI would expect higher customer volumes during weekends than weekdays, whereas the visiting count of the Commercial Banking POI has a significant drop during weekends.
Learning how to distinguish different categories from input prompts could be beneficial in forecasting future visiting volumes.
In addition, another important characteristic of this introduced auxiliary operation is that it does not affect the encoder structure. 
Thus, available pre-trained weights of encoders/decoders can be directly used as initialisation, which is a key enabler for forecasting under the \textit{pre-train and fine-tune paradigm}.

The overall pipeline works in a similar fashion as the typical sequence-to-sequence task. The encoder encodes the input mobility prompt sentences which describe the history mobility observation. The decoder then takes the encoded features as input to generate the predicted sentence tokens.
This process can be described as:
\begin{equation}
    Token_{\text{pred}}=\text{DECODER}(\text{ENCODER}(X)),
\end{equation}
where $\text{ENCODER}(\cdot)$ and $\text{DECODER}(\cdot)$ represent the encoder and the decoder, respectively.
In the proposed \name, we vary and investigate multiple Transformer based encoder structures (details given in Section~\ref{sec:encoder}).
With the help of the self-attention mechanism inside the Transformer, not only the intra-relation of numerical tokens at different time steps (\ie, the mobility data $x^m_{t_1: t_{n}}$) in the input prompts but also the inter-relation between numerical and contextual tokens (\eg, tokens for the date information) could be learned simultaneously to generate future predictions.
GPT-2~\cite{gpt2} is also a Transformers-based model architecture designed to yield the next word in sentences pre-trained with very large corpus of English data (general English text data without any specific mobility data).
It stacks multiple decoder layers (normally 12 - 48 layers), the GPT-2 with 12 decoder layers is selected as the decoder in our \name.
The auto-regressive nature of the GPT-2 decoder makes it a preferred choice for text generation tasks from a given input prompt..
The numerical prediction $\hat{x}_{t_{n+1}}$ can then be detokenized from the generated token $Token_{\text{pred}}$.

Specifically, for the auxiliary classification, after encoding through an encoder (\eg, BERT), the feature embedding of the {\fontfamily{qcr}\selectfont [CLS]} token is passed through a fully connected layer followed by a softmax layer for the POI category classification. 
This token is selected for classification as it is a global token and attends to all the other input tokens that cover the mobility information.

\subsection{Training Objective}
Due to the introduced auxiliary POI category classification, there are two types of losses in \name.
The first loss is a cross-entropy loss denoted by $\mathcal{L}_{\text{CE}}$ for guiding the sequence generation for our main purpose and the second loss $\mathcal{L}_{\text{POI}}$ is the auxiliary category classification loss.
During the training, the proposed \name\ is fine-tuned end-to-end with the loss $\mathcal{L}$ given by: 
\begin{equation}
    \mathcal{L} = \lambda_{\text{CE}}\mathcal{L}_{\text{CE}} + \lambda_{\text{POI}}\mathcal{L}_{\text{POI}} \label{eq:loss},
\end{equation}
where $\lambda_{\text{CE}}$, $\lambda_{\text{POI}}$ are two factors used to combine two losses and the sum of these factors equals 1.

\section{Experiments}\label{sec:exp}
In our experiments, we aim to investigate and answer the following research questions:
\begin{itemize}
    \item \textbf{RQ1}: Compared to the conventional \textit{numerical paradigm} forecasting methods, what is the performance of language based encoder-decoder methods with mobility prompts as input?
    \item \textbf{RQ2}: Under the \name\ pipeline, can we apply different pre-trained language foundation models as encoders, and what is the forecasting performance for each of them?
    \item \textbf{RQ3}: Could we achieve a further mobility forecasting performance gain through the introduced auxiliary POI category classification?
\end{itemize}

\begin{table}[]
\centering
\caption{Pre-trained language models explored in the proposed \name.}
\addtolength{\tabcolsep}{-0.25ex}
\begin{tabular}{c|c|c} \toprule
 & Model & HuggingFace Configuration\\ \midrule
 & BERT & https://huggingface.co/bert-base-uncased \\
Encoder & RoBERTa & https://huggingface.co/roberta-base \\
 & XLNet & https://huggingface.co/xlnet-base-cased \\ \midrule
Decoder & GPT-2 & https://huggingface.co/gpt2 \\ \bottomrule
\end{tabular}
\label{tab:model}
\end{table}

\subsection{Datasets}\label{sec:dataset}
In this study, the datasets used for evaluation are from real-world human mobility data provided by SafeGraph.\footnote{https://docs.safegraph.com/docs/weekly-patterns}
It contains daily visit counting records and the category information of each POI.
To enhance privacy, the provided data is aggregated and anonymised.
We access the raw SafeGraph Weekly Patterns data through SafeGraph Data for Academics\footnote{https://www.safegraph.com/academics}. 
The data from three representative cities (New York City (NYC), Dallas, and Miami) was collected from 2020-06-15 to 2020-11-08 to form three datasets.
These datasets include 479 POIs from 39 categories, 1374 POIs from 65 categories, and 1007 POIs from 51 categories, respectively.
Based on these statistics, we can see that Dallas has the most POIs and categories, while NYC has the least number of POIs/categories.
We split each dataset into training set (70\%), validation set (10\%), and testing set (20\%).
For the input data, the history observation length is defined as 15 days (\ie, $n=15$).
Mobility prompts given in the sentence format are generated based on the templates illustrated in Table\ref{tab:prompt}.
For the comparison purpose, the train/val/test sets in the numerical format are also maintained for each city to evaluate methods in the \textit{numerical paradigm}.

\begin{table*}[]
\centering
\caption{Prediction results of the numerical paradigm based methods and methods using mobility prompts. Both Prompt A and Prompt B are compared.}\label{tab:res}
\addtolength{\tabcolsep}{-0.5ex}
\begin{tabular}{c|c|cc|cc|cc|cc} \toprule
\multirow{2}{*}{Prompt} &\multirow{2}{*}{Model}  & \multicolumn{2}{c|}{NYC} & \multicolumn{2}{c|}{Dallas} & \multicolumn{2}{c|}{Miami} & \multicolumn{2}{c}{Average} \\ \cline{3-10}
 &  & RMSE & MAE & RMSE & MAE & RMSE & MAE & RMSE & MAE \\ \midrule
\multirow{5}{*}{\begin{tabular}[c]{@{}c@{}}N/A\\ Numerical based\end{tabular}} & LR & 9.131 & 5.639 & 24.544 & 6.601 & 13.081 & 6.082 & 15.585 & 6.107 \\
 & GRU & 7.547$\pm$0.098 & 4.550$\pm$0.038 & 23.987$\pm$0.262 & 5.400$\pm$0.016 & 12.125$\pm$0.160 & 5.413$\pm$0.026 & 14.553 & 5.121 \\
 & GRUAtt & 7.704$\pm$0.107 & 4.464$\pm$0.037 & 22.562$\pm$0.433 & 5.276$\pm$0.048 & 11.465$\pm$0.417 & 5.045$\pm$0.107 & 13.910 & 4.928 \\
 & Transformer & 6.714$\pm$0.072 & 4.279$\pm$0.058 & 18.820$\pm$0.278 & 5.166$\pm$0.125 & 10.995$\pm$0.181 & 5.130$\pm$0.117 & 12.176 & 4.858 \\
 \midrule
 
N/A  & Transformer &  6.452$\pm$0.055  &  4.250$\pm$0.057  &  18.796$\pm$0.338  &  5.337$\pm$0.183  &  10.004$\pm$0.022  &  5.053$\pm$0.066  & 11.751 & 4.880 \\
Numerical based & Reformer &  6.645$\pm$0.040  &  4.377$\pm$0.018  &  17.423$\pm$0.200  &  5.518$\pm$0.066  &  10.411$\pm$0.151  &  5.116$\pm$0.046 & 11.493 & 5.004 \\
With Temporal& Informer &  6.279$\pm$0.140  &  4.134$\pm$0.074  &  18.061$\pm$0.205 &  5.441$\pm$0.052  & 9.526$\pm$0.098  &  4.823$\pm$0.043  & 11.289  & 4.799 \\ 
Embedding  & Autoformer & 6.433$\pm$0.103 & 4.323$\pm$0.108 & 18.033$\pm$0.896 & 7.021$\pm$ 0.977 & 9.852$\pm$ 0.731 & 6.321$\pm$0.701 & 11.439 & 5.888 \\\midrule
\multirow{2}{*}{A} & GRUAtt-A & 6.901$\pm$0.212 & 4.290$\pm$0.042 & 19.914$\pm$1.259 & 5.165$\pm$0.067 & 9.964$\pm$0.632 & 5.009$\pm$0.055 & 12.260 & 4.821 \\
 & Transformer-A & 6.657$\pm$0.070 & 4.286$\pm$0.075 & 18.212$\pm$1.422 & 5.036$\pm$0.096 & 9.672$\pm$0.605 & 5.034$\pm$0.105 & 11.514 & 4.785 \\ \midrule
\multirow{2}{*}{B} & GRUAtt-B & 6.887$\pm$0.105 & 4.355$\pm$0.059 & 19.743$\pm$0.884 & 5.212$\pm$ 0.227 & 10.066$\pm$0.520 & 5.124$\pm$0.036 & 12.232 & 4.897 \\
 & Transformer-B & 6.648$\pm$0.190 & 4.273$\pm$0.054 & 18.189$\pm$1.382 & 5.087$\pm$0.023 & 9.563$\pm$0.406 & 4.991$\pm$0.164 & 11.467 & 4.784 \\
 \bottomrule
\end{tabular}

\end{table*}

\subsection{Implementations}

The proposed \name\ follows the \textit{pre-train and fine-tune paradigm} and we leverage the existing pre-trained language models.
The encoders and decoders are configured and initialised with pre-trained weights provided by HuggingFace Models.
We then carry out fine-tuning using the generated mobility.
The pre-trained language models are fine-tuned using mobility prompts with the following implementation settings.
The total epoch number is selected as 50 with early-stopping.
For factors in loss function (Equation~\eqref{eq:loss}), $\lambda_{\text{CE}}=0.8, \lambda_{\text{POI}}=0.2$ is also applied on NYC and Miami datasets, whereas the setting of $\lambda_{\text{CE}}=0.9, \lambda_{\text{POI}}=0.1$ is used for Dallas dataset.
The models are optimised with Adam optimiser with a $5\times10^{-5}$ initial learning rate (weight decay is set to $5\times10^{-4}$).
These parameters are selected based on the performance on the validation set. 
The \textit{ReduceLROnPlateau} decay is adopted with patience 6 and cooldown 2 during fine-tuning.
The experiments are performed with PyTorch on a Linux server equipped with Nvidia V100 GPUs.
For the configurations and pre-trained weights of the pre-trained language models used in our experiments, 
Table~\ref{tab:model} lists the corresponding HuggingFace links.
In the experiments, we used the standard pre-trained weights of these decoder/encoders which can be accessed through the listed links.
Our codes are available at https://github.com/cruiseresearchgroup/AuxMobLCast.

\subsection{Evaluation}
For evaluating the performance of each method, considering we focus on the numerical prediction task, the Root Mean Square Error (RMSE) and the Mean Absolute Error (MAE) are selected as evaluation metrics.
These errors are calculated based on the predicted $\hat{x}_{t_{n+1}}$ and the ground truth $x_{t_{n+1}}$ to measure the closeness of the predicted values.
In our experiments, the average performance and the standard deviation of multiple runs (5 runnings with different random seeds) of each method are reported excluding the basic linear regression (LR) method.

\begin{table*}[]
\centering
\caption{Prediction results of different configurations of our \name\ on three datasets. Prompt C is applied for this part of experiments.}
\begin{tabular}{c|c|cc|cc|cc|cc}\toprule
\multirow{2}{*}{Encoder} &\multirow{2}{*}{Aux} &\multicolumn{2}{c|}{NYC} &\multicolumn{2}{c|}{Dallas} &\multicolumn{2}{c|}{Miami} &\multicolumn{2}{c}{Average} \\ \cline{3-10}
& &RMSE &MAE &RMSE &MAE &RMSE &MAE &RMSE &MAE \\ \hline
\multirow{2}{*}{BERT} &  \checkmark & 6.312$\pm$0.253 & 4.114$\pm$0.038 & 15.304$\pm$0.835 & 5.168$\pm$0.210 & 10.307$\pm$1.698 & 4.804$\pm$0.084 & 10.641 & 4.695\\ \cline{2-10}
& $\times$ & 6.291$\pm$0.010 & 4.144$\pm$0.024 & 18.125$\pm$1.509 & 5.111$\pm$0.096 & 12.197$\pm$1.057 & 4.871$\pm$0.060 & 12.204 & 4.708 \\ \hline
\multirow{2}{*}{RoBERTa} &  \checkmark & 6.277$\pm$0.218 & 4.106$\pm$0.048 & 16.902$\pm$1.621 & 4.964$\pm$0.062 & 10.744$\pm$0.793 & 4.926$\pm$0.127 & 11.307 & 4.665  \\ 
\cline{2-10}
& $\times$ & 6.336$\pm$0.259 & 4.117$\pm$0.049 & 15.821$\pm$1.114 & 5.294$\pm$0.193 & 11.804$\pm$0.652 & 5.228$\pm$0.172 & 11.320 & 4.879\\ \hline
\multirow{2}{*}{XLNet} &  \checkmark & 6.586$\pm$0.177 & 4.289$\pm$0.085 & 16.566$\pm$0.998 & 5.305$\pm$0.094 & 12.683$\pm$1.127 & 5.075$\pm$0.161 & 11.945 & 4.889  \\ \cline{2-10}
& $\times$ & 6.605$\pm$0.253 & 4.223$\pm$0.033 & 15.602$\pm$0.285 & 5.202$\pm$0.123 &13.071$\pm$2.561 & 5.254$\pm$0.059 &  11.759 & 4.893\\ 
\bottomrule
\end{tabular}
\label{tab:aux}
\end{table*}

\subsection{Results and Analysis}

In this part of experiments, we compare the performance of using mobility prompts against the following \textit{numerical paradigm} based time-series forecasting methods:
\begin{itemize}
    \item Basic Linear Regression (LR);
    \item GRU~\cite{chung2014empirical}: one of the most popular variant of Recurrent Neural Networks;
    \item GRUAtt~\cite{BahdanauCB2015}: introducing the attention mechanism into the GRU architecture;
    \item Transformer~\cite{vaswani2017attention}: the vanilla Transformer network with the multi-head self-attention mechanism;
    \item Reformer~\cite{reformer2020}: a variant of Transformer focused on the efficiency. It has been used as a baseline for forecasting in~\cite{zhou2021informer} and \cite{xu2021autoformer}.
    \item Informer~\cite{zhou2021informer}: a specific variant of Transformer that is proposed for long sequence forecasting task.
    \item Autoformer~\cite{xu2021autoformer}: a recent variant of Transformer for predicting temporal sequences.
\end{itemize}
For Transformer based methods (including Transformer, Reformer, Informer, and Autoformer), we consider incorporating temporal information. Specifically, the temporal embeddings (e.g., time-of-day, day-of-week, month-of-year) are injected as the Transformer position embeddings so that the date information is used as input features for prediction.
All the above methods take the numerical sequences as input and generate a number as the prediction.
For methods using natural language based mobility prompts as input/output, we start with two widely used architectures for language sequence-to-sequence tasks, namely, GRUAtt and Transformer. Both methods are in the encoder-decoder structure.
Prompt A and Prompt B defined in Table\ref{tab:prompt} are adopted respectively. To make it clear, we use ``-A'' and ``-B'' to differentiate methods trained with two prompts (\eg, GRUAtt-A, Transformer-B).

\begin{table*}[]
\centering
\caption{Prediction results of different configurations (with or without using temporal date information in the input prompts).}
\begin{tabular}{c|c|cc|cc|cc|cc}\toprule
\multirow{2}{*}{Encoder} &\multirow{2}{*}{Date Info} &\multicolumn{2}{c|}{NYC} &\multicolumn{2}{c|}{Dallas} &\multicolumn{2}{c|}{Miami} &\multicolumn{2}{c}{Average} \\ \cline{3-10}
& &RMSE &MAE &RMSE &MAE &RMSE &MAE &RMSE &MAE \\ \hline
\multirow{2}{*}{BERT} &  \checkmark & 6.312$\pm$0.253 & 4.114$\pm$0.038 & 15.304$\pm$0.835 & 5.168$\pm$0.210 & 10.307$\pm$1.698 & 4.804$\pm$0.084 & 10.641 & 4.695\\ \cline{2-10}
& $\times$ &  6.764$\pm$0.092  &  4.461$\pm$0.040  &  16.633$\pm$0.552  &  5.550$\pm$0.230  & 11.017$\pm$1.348 & 5.331$\pm$0.159 & 11.471  &5.114 \\   \hline
\multirow{2}{*}{RoBERTa} &  \checkmark & 6.277$\pm$0.218 & 4.106$\pm$0.048 & 16.902$\pm$1.621 & 4.964$\pm$0.062 & 10.744$\pm$0.793 & 4.926$\pm$0.127 & 11.307 & 4.665  \\ 
\cline{2-10}
& $\times$ &  6.498$\pm$0.089  &  4.345$\pm$0.036  &  17.091$\pm$0.798  &  5.289$\pm$0.047  & 12.030$\pm$0.355 & 5.353$\pm$0.117 & 11.873  &4.996 \\  \hline
\multirow{2}{*}{XLNet} &  \checkmark & 6.586$\pm$0.177 & 4.289$\pm$0.085 & 16.566$\pm$0.998 & 5.305$\pm$0.094 & 12.683$\pm$1.127 & 5.075$\pm$0.161 & 11.945 & 4.889  \\ \cline{2-10}
& $\times$ &  7.434$\pm$0.371  &  4.944$\pm$0.226  &  17.647$\pm$1.702  & 6.834$\pm$1.540   &  16.605$\pm$3.717  & 7.578$\pm$1.494 & 13.895  & 6.452\\
\bottomrule
\end{tabular}
\label{tab:dateinfo}
\end{table*}

The results of the above methods on the testing set are reported in Table\ref{tab:res}. The last two columns list the average RMSE and MAE across three cities for each method.
In general, for each method, it can be noticed that NYC has the lowest prediction error whereas Dallas has the worst performance.
This observation can be explained by the fact that Dallas is relatively harder to predict due to the largest number of POIs and categories in all three datasets.
For methods under the numerical paradigm, Transformer outperforms other baselines while the linear regression has the worst performance, which is as expected. 
When the temporal information is considered, Transformer with temporal embedding yields a better RMSE performance compared to Transformer without using temporal embedding. Autoformer and Reformer also demonstrate good prediction performance. Among these numerical methods with temporal embedding, Informer achieves the best performance on both RMSE and MAE against all the other numerical based methods.
This shows that the temporal information can be used as a strong cue for forecasting and it should be included the mobility prompts.

Comparing the methods with the same backbone model (GRUAtt vs. GRUAtt-A/B and Transformer vs. Transformer-A/B), we observe that using language based mobility prompts could lead to better prediction performance.
Furthermore, with the help of mobility prompts, the methods even using the vanilla Transformer as backbone (Transformer-A and Transformer-B) could have similar prediction performance (worse performance on average RMSE but better on MAE) as the state-of-the-art numerical based methods (Informer and Autoformer with temporal embedding).
In addition, comparing results using Prompt A and results using Prompt B (the last four rows), it can be seen that the two prompt types have very close performance.
Considering that Prompt B further removes the sensitive POI id information from Prompt A, Prompt B is a preferable prompt type.
The above results and analysis could answer the RQ1 and indicate that using mobility prompts has better prediction performance compared to the basic numerical based forecasting methods and is able to achieve comparable performance as the state-of-the-art numerical based methods with temporal embedding.

\begin{table}[]
\centering
\caption{The comparison of numerical forecasting methods and our \name\ under the zero-shot setting.}
\addtolength{\tabcolsep}{-0.25ex}
\begin{tabular}{c|c|c||cc} \toprule
Training & Test & Method & RMSE & MAE \\ \midrule
 &  & Transformer & 15.867$\pm$0.202 & 5.220$\pm$0.084 \\
 &  & Reformer & 15.488$\pm$0.169 & 5.401$\pm$0.016 \\
 &  & Informer & 16.333$\pm$0.297 & 5.181$\pm$0.067 \\
 & Miami & Autoformer & \textcolor{red}{9.445}$\pm$0.095 & \textcolor{red}{5.020}$\pm$0.049 \\ \cline{3-5}
 &  & XLNet & 18.801$\pm$2.840 & 7.228$\pm$3.960 \\
 &  & BERT & 20.272$\pm$1.432 & 5.949$\pm$0.223 \\
 &  & RoBERTa & 17.834$\pm$0.284 & 5.598$\pm$0.030 \\ \cline{2-5}
NYC &  & Transformer & 31.207$\pm$0.304 & 5.721$\pm$0.098 \\
 &  & Reformer & 30.502$\pm$0.313 & 5.897$\pm$0.022 \\
 &  & Informer & 31.314$\pm$0.827 & 5.615$\pm$0.077 \\
 & Dallas & Autoformer & 19.239$\pm$0.564 & 5.327$\pm$0.065 \\ \cline{3-5}
 &  & XLNet & 21.341$\pm$1.733 & 8.291$\pm$1.008 \\
 &  & BERT & \textcolor{red}{17.396}$\pm$0.995 & 5.472$\pm$0.027 \\
 &  & RoBERTa & 17.415$\pm$0.224 & \textcolor{red}{5.309}$\pm$0.021 \\ \midrule
 &  & Transformer & 6.656$\pm$0.044 & 4.341$\pm$0.023 \\
 &  & Reformer & 7.514$\pm$0.056 & 4.770$\pm$0.035 \\
 &  & Informer & 6.429$\pm$0.074 & 4.236$\pm$0.036 \\
 & NYC & Autoformer & 6.525$\pm$0.065 & 4.432$\pm$0.048 \\ \cline{3-5}
 &  & XLNet & 7.158$\pm$0.178 & 4.304$\pm$0.015 \\
Miami &  & BERT & 6.295$\pm$0.066 & \textcolor{red}{4.204}$\pm$0.019 \\
 &  & RoBERTa & \textcolor{red}{6.289}$\pm$0.061 & 4.209$\pm$0.032 \\ \cline{2-5}
 &  & Transformer & 21.405$\pm$0.373 & 5.316$\pm$0.033 \\
 &  & Reformer & 25.205$\pm$0.832 & 5.723$\pm$0.056 \\
 &  & Informer & 21.688$\pm$0.510 & 5.198$\pm$0.045 \\
 & Dallas & Autoformer & 21.267$\pm$0.990 & 5.350$\pm$0.037 \\ \cline{3-5}
 &  & XLNet & 16.747$\pm$0.150 & \textcolor{red}{5.149}$\pm$0.019 \\
 &  & BERT & \textcolor{red}{15.546}$\pm$0.241 & 5.723$\pm$0.224 \\
 &  & RoBERTa & 20.920$\pm$1.245 & 5.202$\pm$0.048 \\ \midrule
 &  & Transformer & 6.733$\pm$0.753 & 4.447$\pm$0.066 \\
 &  & Reformer & 7.556$\pm$0.036 & 4.823$\pm$0.023 \\
 &  & Informer & 6.766$\pm$0.078 & 4.497$\pm$0.075 \\
 & NYC & Autoformer & 6.939$\pm$0.204 & 4.855$\pm$0.167 \\ \cline{3-5}
 &  & XLNet & 7.202$\pm$0.371 & 4.702$\pm$0.225 \\
 &  & BERT & \textcolor{red}{6.231}$\pm$0.066 & \textcolor{red}{4.162}$\pm$0.017 \\
Dallas &  & RoBERTa & 6.291$\pm$0.144 & 4.249$\pm$0.090 \\ \cline{2-5}
 &  & Transformer & 10.904$\pm$0.129 & 5.995$\pm$0.037 \\
 &  & Reformer & 11.259$\pm$0.715 & 5.287$\pm$0.059 \\
 &  & Informer & \textcolor{red}{9.657}$\pm$0.422 & \textcolor{red}{5.076}$\pm$0.043 \\
 & Miami & Autoformer & 10.321$\pm$0.665 & 5.457$\pm$0.128 \\ \cline{3-5}
 &  & XLNet & 15.801$\pm$2.490 & 5.771$\pm$0.291 \\
 &  & BERT & 14.014$\pm$0.741 & 5.342$\pm$0.055 \\
 &  & RoBERTa & 16.031$\pm$0.626 & 5.330$\pm$0.123 \\  
 \bottomrule
\end{tabular}
\label{tab:zeroshot}
\end{table}


\subsection{Performance of \name\ and Ablation Study}\label{sec:encoder}
In this section, we focus on answering RQ2 and RQ3.
In the proposed \name, autoregressive language model GPT-2 is selected as the decoder and we explore the performance of using the following language foundation models as the encoder: 
\begin{itemize}
    \item BERT~\citep{bert}: bert-base-uncased, 12-layer, 768-hidden, 12-heads, 110M parameters.
    \item RoBERTa~\citep{liu2019roberta}: roberta-base, 12-layer, 768-hidden, 12-heads, 125M parameters.
    \item XLNet~\citep{yang2019xlnet}: xlnet-base-cased, 12-layer, 768-hidden, 12-heads, 110M parameters.
\end{itemize}
The details of these models and corresponding pre-trained weights are available through the links given in Table~\ref{tab:model}.

The prediction results of using different encoders on three datasets are listed in Table~\ref{tab:aux}.
In this table, a $\checkmark$ under the Aux column indicates that the auxiliary POI category classification is enabled, while a $\times$ means that the auxiliary classification is removed and \name\ is the basic encoder-decoder structure.
In this part of the experiments, Prompt C given in Table\ref{tab:prompt} is applied as inputs and output targets.
When the auxiliary classification is disabled (rows marked with $\times$), the POI category target is also dropped from the Prompt C and only the mobility target part is kept during the fine-tuning process.
\subsubsection{Different Encoders}

If we jointly compare the results (rows with $\checkmark$) reported in Table\ref{tab:aux} with the top performers in Table\ref{tab:res}, we observe that using BERT as the encoder in our \name\ outperforms the performance of Informer and Transformer-B on average.
Although the results of BERT and RoBERTa on Miami are not the best, they achieve significant performance improvements on Dallas, compared to all the baselines in Table\ref{tab:res}.
More specifically, BERT is almost on par with RoBERTa on NYC dataset, while using BERT further reduces the RMSE by 9.5\% on Dallas.
Compared to BERT and RoBERTa, using XLNet seems not optimal under our \name\ pipeline. Its average forecasting performance is slightly worse than Transformer-B.
However, using XLNet as the encoder still has a good forecasting performance on the most challenging Dallas dataset.
These results suggest that our \name\ pipeline can be adapted to multiple language foundation models and BERT is the most suitable encoder structure.
It also justifies our hypothesis that applying pre-trained language foundation models with mobility prompts could be a new direction for addressing the human mobility forecasting task.


\subsubsection{With or Without Auxiliary Classification}

Table\ref{tab:aux} also reports the ablation study results of comparing the row with a $\checkmark$ against the row with a $\times$ under each encoder setting.
On average, introducing the auxiliary POI category classification task brings performance gain for the frameworks using BERT or RoBERTa as the encoder.
For the configuration using XLNet as the encoder, enabling the auxiliary task results in a better forecasting performance on the Miami dataset but has a worse performance (both RMSE and MAE) on the Dallas dataset. 
Therefore, XLNet is not the optimal encoder under the proposed \name\ pipeline.
For the RoBERTa encoder setting, we witness a larger RMSE but a smaller MAE performance on using the auxiliary task on the Dallas dataset.
From the table, we can also notice that when BERT is chosen as the encoder, using the auxiliary task outperforms the setting without the auxiliary part by a relatively large margin on the RMSE of Dallas and Miami datasets.
Based on the above analysis, although the auxiliary POI category classification does not reduce the forecasting errors for every encoder structure, it still yields substantial improvements for the BERT encoder setting.
This further confirms that BERT is a favourable encoder selection for \name.

\subsubsection{With or Without Date Information in Prompts}

Based on the previous experiments (results given in Table~\ref{tab:res}), we have noticed that the temporal embedding could contribute to prediction performance gain for the numerical based forecasting methods. 
In this part of the experiment, we would also like to investigate the importance of temporal information (\ie, the date information of the mobility data) under the proposed \name\ framework.
To this end, we further evaluate the prompts without including the temporal date information.
By removing the temporal information (the blue parts in Table~\ref{tab:prompt}), the input prompt is then simplified as: {\fontfamily{qcr}\selectfont there were {11, 11, 10, 12, 9, 12, 6, 13, 10, 15, 16, 8, 8, 13, 19} people visiting POI \\ on each day.}\footnote{Using the same example as shown in Table~\ref{tab:prompt}.}

Based on this simplified prompt, Table~\ref{tab:dateinfo} lists and compares the performance of our~\name\ under the settings of with or without temporal date information.
Similar to previous experiments, we also evaluate \name\ with different pre-trained language models (\ie, BERT, RoBERTa, and XLNet).
From the table, it can be observed that the performance of all three models have decreased after removing the date information.
More specifically, the performance reduction of using RoBERTa is smaller than the configurations of using BERT and XLNet.
Without the date information in the input prompts, it is difficult for language models to capture the relationships (\eg, weekly patterns) between numerical tokens and temporal contextual tokens. 
The lacking of modelling such relationships could explain the performance reduction.

\subsection{Zero-shot Performance}
To further investigate the performance of \name, we conduct an experiment under the zero-shot setting. Specifically, we train each method on one dataset and test the trained model on the test set of the rest two datasets (\eg, train on NYC, test on Miami and Dallas).
The comparison results of the state-of-the-art numerical based forecasting methods (including Transformer, Reformer, Informer, and Autoformer) and our \name\ (using XLNet, BERT, RoBERTa as encoders) are reported in Table~\ref{tab:zeroshot}. For numerical based methods, the temporal embeddings are enabled to consider the temporal information. The best performer under each training/test configuration has been shown in red in the table.
From this table, it can be observed that \name\ achieves 4 top performers in a total of 6 configurations. The numerical based methods have better performance when the test set is Miami (for both training with NYC and training with Dallas situations).
Although using the language foundation models in our \name\ does not demonstrate superior performance for all settings, it still shows that \name\ has a relatively good and promising generalisation ability.
With further research, we believe that using pre-trained language models would achieve better and more robust performance in the task of human mobility forecasting.

\section{Discussion}\label{sec:conclusion}
In this work, we studied the application of language models especially pre-trained language foundation models for mining temporal sequential patterns to predict human mobility.
To this end, mobility prompts are applied to transform the numerical mobility data into sentences as inputs and output targets.
To utilise pre-trained language models, we also propose the \name\ based on the encoder-decoder architecture, in which an auxiliary task is explicitly introduced for classifying POI categories.
The results show that addressing sequential numerical data prediction through mobility prompts and applying pre-trained language foundation models (\eg, BERT) under the proposed \name\ pipeline could improve the forecasting performance.

\smallskip
\noindent\textbf{Why It Works.}

In recent years, Transformer and its variants have emerged as powerful models in addressing NLP tasks. Through the inherent attention mechanisms, Transformers can well discovering the latent relationship of the input sequence tokens. 
More recently, we have also witnessed the booming of Vision Transformers for processing images or videos. Through the patching transformation, images are divided into patch sequences to fit the Transformer structure.
Thus, for the time-series sequences (\eg, human mobility data), we believe pre-trained language Transformers could also suitable in handling time-series data with proper transformation from the raw numerical data to language sequences (\ie, the mobility prompts).
Through the mobility prompts, the intra-relation of numerical values at different time steps as well as the inter-relation between numerical values and contextual information (\eg, temporal date information) would be better modelled simultaneously by Transformers, which leads to good forecasting performance.

\smallskip
\noindent\textbf{Broader Impact.}

The findings of this research provide visionary ideas and novel insights for human mobility forecasting tasks.
The human mobility forecasting task analysed in this paper shows a glimpse of the potential applications in the direction of applying pre-trained language models with prompts.
How to leverage pre-trained language models for various numerical forecasting applications (\eg, weather forecasting, demand forecasting) could lead to new research directions.
Although the improvement gain demonstrated in this paper is not very large, we believe that with further and deeper research, using pre-trained language models could be a new trend for the spatio-temporal and human mobility forecasting areas.
Further, we hope this work would also facilitate the related research of multimedia assistant systems (\eg, integrating spatio-temporal forecasting ability with Siri or Cortana).

\smallskip
\noindent\textbf{Future Work.}

As this is the first research that leverages pre-trained language models for discovering temporal sequential patterns in mobility forecasting tasks, there are still several limitations. One limitation of this study is about the mobility prompt generation.
In the future, we plan to fully investigate mobility prompts based on the recent prompt learning techniques.
An automatic approach for transforming diverse sequential numerical behaviour data, as well as various types of time-series data, will be beneficial in exploring the forecasting ability of pre-trained language models.
In addition, how to explore pre-trained language models for multi-variate time-series data forecasting could be another interesting future direction.

\begin{acks}
This work was supported by Australian Research Council (ARC) Discovery Project \textit{DP190101485}. This paper is also a contribution to the IEA EBC Annex 79. 
\end{acks}

\bibliographystyle{ACM-Reference-Format}
\bibliography{main}


\begin{thebibliography}{23}


\ifx \showCODEN    \undefined \def \showCODEN     #1{\unskip}     \fi
\ifx \showDOI      \undefined \def \showDOI       #1{#1}\fi
\ifx \showISBNx    \undefined \def \showISBNx     #1{\unskip}     \fi
\ifx \showISBNxiii \undefined \def \showISBNxiii  #1{\unskip}     \fi
\ifx \showISSN     \undefined \def \showISSN      #1{\unskip}     \fi
\ifx \showLCCN     \undefined \def \showLCCN      #1{\unskip}     \fi
\ifx \shownote     \undefined \def \shownote      #1{#1}          \fi
\ifx \showarticletitle \undefined \def \showarticletitle #1{#1}   \fi
\ifx \showURL      \undefined \def \showURL       {\relax}        \fi
\providecommand\bibfield[2]{#2}
\providecommand\bibinfo[2]{#2}
\providecommand\natexlab[1]{#1}
\providecommand\showeprint[2][]{arXiv:#2}

\bibitem[Baevski and Mohamed(2020)]%
        {baevski2020effectiveness}
\bibfield{author}{\bibinfo{person}{Alexei Baevski} {and}
  \bibinfo{person}{Abdelrahman Mohamed}.} \bibinfo{year}{2020}\natexlab{}.
\newblock \showarticletitle{Effectiveness of self-supervised pre-training for
  asr}. In \bibinfo{booktitle}{\emph{ICASSP 2020-2020 IEEE International
  Conference on Acoustics, Speech and Signal Processing (ICASSP)}}. IEEE,
  \bibinfo{pages}{7694--7698}.
\newblock


\bibitem[Bahdanau et~al\mbox{.}(2015)]%
        {BahdanauCB2015}
\bibfield{author}{\bibinfo{person}{Dzmitry Bahdanau},
  \bibinfo{person}{Kyunghyun Cho}, {and} \bibinfo{person}{Yoshua Bengio}.}
  \bibinfo{year}{2015}\natexlab{}.
\newblock \showarticletitle{Neural Machine Translation by Jointly Learning to
  Align and Translate}. In \bibinfo{booktitle}{\emph{3rd International
  Conference on Learning Representations, {ICLR} 2015, San Diego, CA, USA, May
  7-9, 2015, Conference Track Proceedings}},
  \bibfield{editor}{\bibinfo{person}{Yoshua Bengio} {and} \bibinfo{person}{Yann
  LeCun}} (Eds.).
\newblock


\bibitem[Berg-Kirkpatrick and Spokoyny(2020)]%
        {berg2020empirical}
\bibfield{author}{\bibinfo{person}{Taylor Berg-Kirkpatrick} {and}
  \bibinfo{person}{Daniel Spokoyny}.} \bibinfo{year}{2020}\natexlab{}.
\newblock \showarticletitle{An empirical investigation of contextualized number
  prediction}. In \bibinfo{booktitle}{\emph{Proceedings of the 2020 Conference
  on Empirical Methods in Natural Language Processing (EMNLP)}}.
  \bibinfo{pages}{4754--4764}.
\newblock


\bibitem[Bommasani et~al\mbox{.}(2021)]%
        {bommasani2021opportunities}
\bibfield{author}{\bibinfo{person}{Rishi Bommasani}, \bibinfo{person}{Drew~A
  Hudson}, \bibinfo{person}{Ehsan Adeli}, \bibinfo{person}{Russ Altman},
  \bibinfo{person}{Simran Arora}, \bibinfo{person}{Sydney von Arx},
  \bibinfo{person}{Michael~S Bernstein}, \bibinfo{person}{Jeannette Bohg},
  \bibinfo{person}{Antoine Bosselut}, \bibinfo{person}{Emma Brunskill},
  {et~al\mbox{.}}} \bibinfo{year}{2021}\natexlab{}.
\newblock \showarticletitle{On the Opportunities and Risks of Foundation
  Models}.
\newblock \bibinfo{journal}{\emph{arXiv preprint arXiv:2108.07258}}
  (\bibinfo{year}{2021}).
\newblock


\bibitem[Chung et~al\mbox{.}(2014)]%
        {chung2014empirical}
\bibfield{author}{\bibinfo{person}{Junyoung Chung}, \bibinfo{person}{Caglar
  Gulcehre}, \bibinfo{person}{KyungHyun Cho}, {and} \bibinfo{person}{Yoshua
  Bengio}.} \bibinfo{year}{2014}\natexlab{}.
\newblock \showarticletitle{{Empirical Evaluation of Gated Recurrent Neural
  Networks on Sequence Modeling}}.
\newblock \bibinfo{journal}{\emph{arXiv preprint arXiv:1412.3555}}
  (\bibinfo{year}{2014}).
\newblock


\bibitem[Devlin et~al\mbox{.}(2019)]%
        {bert}
\bibfield{author}{\bibinfo{person}{Jacob Devlin}, \bibinfo{person}{Ming{-}Wei
  Chang}, \bibinfo{person}{Kenton Lee}, {and} \bibinfo{person}{Kristina
  Toutanova}.} \bibinfo{year}{2019}\natexlab{}.
\newblock \showarticletitle{{BERT:} Pre-training of Deep Bidirectional
  Transformers for Language Understanding}. In
  \bibinfo{booktitle}{\emph{Proceedings of the 2019 Conference of the North
  American Chapter of the Association for Computational Linguistics: Human
  Language Technologies, {NAACL-HLT}}}, \bibfield{editor}{\bibinfo{person}{Jill
  Burstein}, \bibinfo{person}{Christy Doran}, {and} \bibinfo{person}{Thamar
  Solorio}} (Eds.). \bibinfo{publisher}{Association for Computational
  Linguistics}, \bibinfo{pages}{4171--4186}.
\newblock


\bibitem[Feng et~al\mbox{.}(2018)]%
        {feng2018deepmove}
\bibfield{author}{\bibinfo{person}{Jie Feng}, \bibinfo{person}{Yong Li},
  \bibinfo{person}{Chao Zhang}, \bibinfo{person}{Funing Sun},
  \bibinfo{person}{Fanchao Meng}, \bibinfo{person}{Ang Guo}, {and}
  \bibinfo{person}{Depeng Jin}.} \bibinfo{year}{2018}\natexlab{}.
\newblock \showarticletitle{Deepmove: Predicting human mobility with
  attentional recurrent networks}. In \bibinfo{booktitle}{\emph{Proceedings of
  the 2018 world wide web conference}}. \bibinfo{pages}{1459--1468}.
\newblock


\bibitem[Hochreiter and Schmidhuber(1997)]%
        {hochreiter1997long}
\bibfield{author}{\bibinfo{person}{Sepp Hochreiter} {and}
  \bibinfo{person}{J{\"u}rgen Schmidhuber}.} \bibinfo{year}{1997}\natexlab{}.
\newblock \showarticletitle{Long Short-term Memory}.
\newblock \bibinfo{journal}{\emph{Neural computation}} \bibinfo{volume}{9},
  \bibinfo{number}{8} (\bibinfo{year}{1997}), \bibinfo{pages}{1735--1780}.
\newblock


\bibitem[Kitaev et~al\mbox{.}(2020)]%
        {reformer2020}
\bibfield{author}{\bibinfo{person}{Nikita Kitaev}, \bibinfo{person}{Lukasz
  Kaiser}, {and} \bibinfo{person}{Anselm Levskaya}.}
  \bibinfo{year}{2020}\natexlab{}.
\newblock \showarticletitle{Reformer: The Efficient Transformer}. In
  \bibinfo{booktitle}{\emph{8th International Conference on Learning
  Representations, {ICLR} 2020, Addis Ababa, Ethiopia, April 26-30, 2020}}.
  \bibinfo{publisher}{OpenReview.net}.
\newblock


\bibitem[Li et~al\mbox{.}(2019)]%
        {li2019enhancing}
\bibfield{author}{\bibinfo{person}{Shiyang Li}, \bibinfo{person}{Xiaoyong Jin},
  \bibinfo{person}{Yao Xuan}, \bibinfo{person}{Xiyou Zhou},
  \bibinfo{person}{Wenhu Chen}, \bibinfo{person}{Yu-Xiang Wang}, {and}
  \bibinfo{person}{Xifeng Yan}.} \bibinfo{year}{2019}\natexlab{}.
\newblock \showarticletitle{Enhancing the locality and breaking the memory
  bottleneck of transformer on time series forecasting}.
\newblock \bibinfo{journal}{\emph{Advances in Neural Information Processing
  Systems}}  \bibinfo{volume}{32} (\bibinfo{year}{2019}),
  \bibinfo{pages}{5243--5253}.
\newblock


\bibitem[Liu et~al\mbox{.}(2016)]%
        {liu2016predicting}
\bibfield{author}{\bibinfo{person}{Qiang Liu}, \bibinfo{person}{Shu Wu},
  \bibinfo{person}{Liang Wang}, {and} \bibinfo{person}{Tieniu Tan}.}
  \bibinfo{year}{2016}\natexlab{}.
\newblock \showarticletitle{Predicting the next location: A recurrent model
  with spatial and temporal contexts}. In \bibinfo{booktitle}{\emph{Thirtieth
  AAAI conference on artificial intelligence}}.
\newblock


\bibitem[Liu et~al\mbox{.}(2019)]%
        {liu2019roberta}
\bibfield{author}{\bibinfo{person}{Yinhan Liu}, \bibinfo{person}{Myle Ott},
  \bibinfo{person}{Naman Goyal}, \bibinfo{person}{Jingfei Du},
  \bibinfo{person}{Mandar Joshi}, \bibinfo{person}{Danqi Chen},
  \bibinfo{person}{Omer Levy}, \bibinfo{person}{Mike Lewis},
  \bibinfo{person}{Luke Zettlemoyer}, {and} \bibinfo{person}{Veselin
  Stoyanov}.} \bibinfo{year}{2019}\natexlab{}.
\newblock \showarticletitle{Roberta: A robustly optimized bert pretraining
  approach}.
\newblock \bibinfo{journal}{\emph{arXiv preprint arXiv:1907.11692}}
  (\bibinfo{year}{2019}).
\newblock


\bibitem[Radford et~al\mbox{.}(2021)]%
        {clip}
\bibfield{author}{\bibinfo{person}{Alec Radford}, \bibinfo{person}{Jong~Wook
  Kim}, \bibinfo{person}{Chris Hallacy}, \bibinfo{person}{Aditya Ramesh},
  \bibinfo{person}{Gabriel Goh}, \bibinfo{person}{Sandhini Agarwal},
  \bibinfo{person}{Girish Sastry}, \bibinfo{person}{Amanda Askell},
  \bibinfo{person}{Pamela Mishkin}, \bibinfo{person}{Jack Clark},
  \bibinfo{person}{Gretchen Krueger}, {and} \bibinfo{person}{Ilya Sutskever}.}
  \bibinfo{year}{2021}\natexlab{}.
\newblock \showarticletitle{Learning Transferable Visual Models From Natural
  Language Supervision}. In \bibinfo{booktitle}{\emph{ICML}}
  \emph{(\bibinfo{series}{Proceedings of Machine Learning Research},
  Vol.~\bibinfo{volume}{139})}, \bibfield{editor}{\bibinfo{person}{Marina
  Meila} {and} \bibinfo{person}{Tong Zhang}} (Eds.).
  \bibinfo{publisher}{{PMLR}}, \bibinfo{pages}{8748--8763}.
\newblock


\bibitem[Radford et~al\mbox{.}(2019)]%
        {gpt2}
\bibfield{author}{\bibinfo{person}{Alec Radford}, \bibinfo{person}{Jeffrey Wu},
  \bibinfo{person}{Rewon Child}, \bibinfo{person}{David Luan},
  \bibinfo{person}{Dario Amodei}, \bibinfo{person}{Ilya Sutskever},
  {et~al\mbox{.}}} \bibinfo{year}{2019}\natexlab{}.
\newblock \showarticletitle{Language models are unsupervised multitask
  learners}.
\newblock \bibinfo{journal}{\emph{OpenAI blog}} \bibinfo{volume}{1},
  \bibinfo{number}{8} (\bibinfo{year}{2019}), \bibinfo{pages}{9}.
\newblock


\bibitem[Spithourakis and Riedel(2018)]%
        {spithourakis2018numeracy}
\bibfield{author}{\bibinfo{person}{Georgios Spithourakis} {and}
  \bibinfo{person}{Sebastian Riedel}.} \bibinfo{year}{2018}\natexlab{}.
\newblock \showarticletitle{Numeracy for Language Models: Evaluating and
  Improving their Ability to Predict Numbers}. In
  \bibinfo{booktitle}{\emph{Proceedings of the 56th Annual Meeting of the
  Association for Computational Linguistics (Volume 1: Long Papers)}}.
  \bibinfo{pages}{2104--2115}.
\newblock


\bibitem[Vaswani et~al\mbox{.}(2017)]%
        {vaswani2017attention}
\bibfield{author}{\bibinfo{person}{Ashish Vaswani}, \bibinfo{person}{Noam
  Shazeer}, \bibinfo{person}{Niki Parmar}, \bibinfo{person}{Jakob Uszkoreit},
  \bibinfo{person}{Llion Jones}, \bibinfo{person}{Aidan~N Gomez},
  \bibinfo{person}{{\L}ukasz Kaiser}, {and} \bibinfo{person}{Illia
  Polosukhin}.} \bibinfo{year}{2017}\natexlab{}.
\newblock \showarticletitle{Attention is all you need}. In
  \bibinfo{booktitle}{\emph{Advances in neural information processing
  systems}}. \bibinfo{pages}{5998--6008}.
\newblock


\bibitem[Wu et~al\mbox{.}(2020)]%
        {wu2020hierarchically}
\bibfield{author}{\bibinfo{person}{Xian Wu}, \bibinfo{person}{Chao Huang},
  \bibinfo{person}{Chuxu Zhang}, {and} \bibinfo{person}{Nitesh~V Chawla}.}
  \bibinfo{year}{2020}\natexlab{}.
\newblock \showarticletitle{Hierarchically structured transformer networks for
  fine-grained spatial event forecasting}. In
  \bibinfo{booktitle}{\emph{Proceedings of The Web Conference 2020}}.
  \bibinfo{pages}{2320--2330}.
\newblock


\bibitem[Xu et~al\mbox{.}(2021)]%
        {xu2021autoformer}
\bibfield{author}{\bibinfo{person}{Jiehui Xu}, \bibinfo{person}{Jianmin Wang},
  \bibinfo{person}{Mingsheng Long}, {et~al\mbox{.}}}
  \bibinfo{year}{2021}\natexlab{}.
\newblock \showarticletitle{Autoformer: Decomposition transformers with
  auto-correlation for long-term series forecasting}.
\newblock \bibinfo{journal}{\emph{Advances in Neural Information Processing
  Systems}}  \bibinfo{volume}{34} (\bibinfo{year}{2021}).
\newblock


\bibitem[Xue et~al\mbox{.}(2021)]%
        {xue2021mobtcast}
\bibfield{author}{\bibinfo{person}{Hao Xue}, \bibinfo{person}{Flora Salim},
  \bibinfo{person}{Yongli Ren}, {and} \bibinfo{person}{Nuria Oliver}.}
  \bibinfo{year}{2021}\natexlab{}.
\newblock \showarticletitle{MobTCast: Leveraging Auxiliary Trajectory
  Forecasting for Human Mobility Prediction}.
\newblock \bibinfo{journal}{\emph{Advances in Neural Information Processing
  Systems}}  \bibinfo{volume}{34} (\bibinfo{year}{2021}).
\newblock


\bibitem[Xue and Salim(2021)]%
        {termcast}
\bibfield{author}{\bibinfo{person}{Hao Xue} {and} \bibinfo{person}{Flora~D.
  Salim}.} \bibinfo{year}{2021}\natexlab{}.
\newblock \showarticletitle{TERMCast: Temporal Relation Modeling for Effective
  Urban Flow Forecasting}. In \bibinfo{booktitle}{\emph{Advances in Knowledge
  Discovery and Data Mining - 25th Pacific-Asia Conference, {PAKDD} 2021,
  Virtual Event, May 11-14, 2021,}}, Vol.~\bibinfo{volume}{12712}.
  \bibinfo{publisher}{Springer}, \bibinfo{pages}{741--753}.
\newblock


\bibitem[Yang et~al\mbox{.}(2019)]%
        {yang2019xlnet}
\bibfield{author}{\bibinfo{person}{Zhilin Yang}, \bibinfo{person}{Zihang Dai},
  \bibinfo{person}{Yiming Yang}, \bibinfo{person}{Jaime Carbonell},
  \bibinfo{person}{Russ~R Salakhutdinov}, {and} \bibinfo{person}{Quoc~V Le}.}
  \bibinfo{year}{2019}\natexlab{}.
\newblock \showarticletitle{Xlnet: Generalized autoregressive pretraining for
  language understanding}.
\newblock \bibinfo{journal}{\emph{Advances in neural information processing
  systems}}  \bibinfo{volume}{32} (\bibinfo{year}{2019}).
\newblock


\bibitem[Zhou et~al\mbox{.}(2021b)]%
        {zhou2021informer}
\bibfield{author}{\bibinfo{person}{Haoyi Zhou}, \bibinfo{person}{Shanghang
  Zhang}, \bibinfo{person}{Jieqi Peng}, \bibinfo{person}{Shuai Zhang},
  \bibinfo{person}{Jianxin Li}, \bibinfo{person}{Hui Xiong}, {and}
  \bibinfo{person}{Wancai Zhang}.} \bibinfo{year}{2021}\natexlab{b}.
\newblock \showarticletitle{Informer: Beyond efficient transformer for long
  sequence time-series forecasting}. In \bibinfo{booktitle}{\emph{Proceedings
  of AAAI}}.
\newblock


\bibitem[Zhou et~al\mbox{.}(2021a)]%
        {zhou2021learning}
\bibfield{author}{\bibinfo{person}{Kaiyang Zhou}, \bibinfo{person}{Jingkang
  Yang}, \bibinfo{person}{Chen~Change Loy}, {and} \bibinfo{person}{Ziwei Liu}.}
  \bibinfo{year}{2021}\natexlab{a}.
\newblock \showarticletitle{Learning to prompt for vision-language models}.
\newblock \bibinfo{journal}{\emph{arXiv preprint arXiv:2109.01134}}
  (\bibinfo{year}{2021}).
\newblock


\end{thebibliography}

\appendix

\end{document}